%% file: main.tex
\documentclass[akbc,twoside,11pt,lettersize]{article}
\usepackage{akbc}
\akbcheading
\ShortHeadings{Entity-Centric Query Refinement}{Wadden, Gupta, Lee \& Toutanova}
\finalcopy 

\usepackage{ntheorem}
\usepackage{amssymb}
\usepackage{amsmath}
\usepackage{bbm}
\usepackage{graphicx}
\usepackage{xcolor,soul,lipsum}
\usepackage{hyperref}
\usepackage{xspace}
\usepackage{tabularx}
\usepackage{booktabs}
\usepackage{array,ragged2e}
\usepackage{multirow}
\usepackage{multicol}
\usepackage{pifont}
\usepackage{algpseudocode}
\usepackage{algorithm}
\hypersetup{colorlinks=true}
\usepackage{subcaption}
\usepackage[colorinlistoftodos,prependcaption,textsize=tiny]{todonotes}
\usepackage{enumitem}
\setlist{parskip=0.5pt}
\setlist{topsep=0.5pt}
\usepackage{wrapfig}

\input{gww-chars.tex}

\input{commands.tex}
\begin{document}

\title{Entity-Centric Query Refinement}

\author{\name David Wadden\addr\textsuperscript{1}\thanks{\vspace{-1em}Work primarily completed while interning at Google.} \email dwadden@cs.washington.edu \\
  \name Nikita Gupta\addr\textsuperscript{2} \email gnikita@google.com \\
  \name Kenton Lee\addr\textsuperscript{2} \email kentonl@google.com \\
  \name Kristina Toutanova\addr\textsuperscript{2} \email kristout@google.com \\
  \addr \textsuperscript{1}Paul G. Allen School of Computer Science \& Engineering, University of Washington \\
  \textsuperscript{2}Google Inc.
}


\maketitle


\input{sec_abstract.tex}
\input{sec_intro.tex}
\input{sec_task_definition.tex}

\input{sec_method.tex}
\input{sec_dataset.tex}

\input{sec_generation.tex}
\input{sec_related_work.tex}
\input{sec_conclusion.tex}
\input{sec_acknowledgments.tex}

\bibliography{refs}
\bibliographystyle{plainnat}

\appendix

\input{appx_qresp_tacl.tex}
\input{appx_dataset.tex}

\input{appx_annotations.tex}

\input{appx_stats.tex}
\input{appx_generation.tex}

\input{appx_eval_queries.tex}

\input{appx_annotation_guide.tex}

\end{document}

%% file: gww-chars.tex
\newcommand{\cA}{{\mathcal{A}}}

\newcommand{\cC}{{\mathcal{C}}}

\newcommand{\cD}{{\mathcal{D}}}

\newcommand{\cM}{{\mathcal{M}}}

\newcommand{\cR}{{\mathcal{R}}}
\newcommand{\fR}{{\mathfrak{R}}}

\newcommand{\cS}{{\mathcal{S}}}



%% file: commands.tex
\newcommand{\sysname}{\texttt{QRESP}\xspace}

\newcommand{\ind}{\mathbbm{1}}
\newcommand{\sumJ}{\sum_{j=1}^{n}}
\newcommand{\sumI}{\sum_{i=1}^k}
\newcommand{\aij}{a_{ij}}

\newcommand{\cRa}{\cR_{\text{A}}}
\newcommand{\cRb}{\cR_{\text{B}}}
\newcommand{\cCFilter}{\cC_{\text{Filter}}}

\newcommand{\Dsys}{\cD_{\sysname}}
\newcommand{\Rsys}{\cR_{\sysname}}
\newcommand{\Drandom}{\cD_{\text{Random}}}
\newcommand{\Dfilter}{\cD_{\text{Random-F}}}
\newcommand{\Rfilter}{\cR_{\text{Random-F}}}
\newcommand{\Rrandom}{\cR_{\text{Random}}}
\newcommand{\Msys}{\cM_{\sysname}}
\newcommand{\Mfilter}{\cM_{\text{Random-F}}}
\newcommand{\Mrandom}{\cM_{\text{Random}}}

\newcommand{\Mseparate}{\cM_{\text{Separate}}}

\newcommand{\filtername}{Random-F}
\newcommand{\nmax}{n_{max}}

\newcolumntype{R}[1]{>{\RaggedLeft\arraybackslash}p{#1}}
\newcolumntype{L}[1]{>{\RaggedRight\arraybackslash}p{#1}}

\newcommand\tworows[1]{\multirow{2}{*}{\shortstack[l]{#1}}}

\newcommand\fourrows[1]{\multirow{4}{*}{\shortstack[l]{#1}}}

\definecolor{cmark}{HTML}{008B72}
\definecolor{xmark}{HTML}{B7321C}
\newcommand{\cmark}{\textcolor{cmark}{\ding{51}\xspace}}
\newcommand{\xmark}{\textcolor{xmark}{\ding{55}\xspace}}

\newcommand{\numQueriesUnfiltered}{17,598\xspace}  
\newcommand{\numQueriesFiltered}{8,958\xspace}  
\newcommand{\numQueriesDev}{282\xspace}   
\newcommand{\minAnswersDev}{200\xspace}   
\newcommand{\minrefinementsDev}{15\xspace}   
\newcommand{\numQueriesYagoEval}{105\xspace}   
\newcommand{\numQueriesNQEval}{93\xspace}   

\newcommand{\nqtrec}{NQ+TREC\xspace}   

\DeclareMathOperator*{\argmin}{arg\,min}

\newcounter{kt}

\newcounter{djwc}

%% file: sec_abstract.tex

\vspace{-1em}

\begin{abstract}
    We introduce the task of entity-centric query refinement. Given an input query whose answer is a (potentially large) collection of entities, the task output is a small set of query \emph{refinements} meant to assist the user in efficient domain exploration and entity discovery. We propose a method to create a training dataset for this task. For a given input query, we use an existing knowledge base taxonomy as a source of candidate query refinements, and choose a final set of refinements from among these candidates using a search procedure designed to partition the set of entities answering the input query. We demonstrate that our approach identifies refinement sets which human annotators judge to be interesting, comprehensive, and non-redundant. In addition, we find that a text generation model trained on our newly-constructed dataset is able to offer refinements for novel queries not covered by an existing taxonomy. Our code and data are available at \url{https://github.com/google-research/language/tree/master/language/qresp}.
\end{abstract}



%% file: sec_intro.tex
\section{Introduction} \label{sec:intro}

During interactive search, the system user may issue queries that are under-specified, ambiguous, or open-ended. For instance, a user interested in finding a new movie might search for ``Action films'', or a computer vision researcher interested in learning more about NLP might search for ``Pretrained NLP models''. These forms of interaction are examples of \emph{exploratory search} \cite{Marchionini2006Exploratory,White2009ExploratorySB}.

We focus specifically on queries whose answer is a list of entities, known as \emph{list-intent} queries. For example, ``Rush Hour'' is one of the thousands of answers to the query ``Action films''. List-intent queries are common, comprising 10\% of all web searches \cite{Chakrabarti2020TableQnAAL}. However, simply displaying the answer to such a query (e.g. a list of all action films) is more likely to cause confusion than to satisfy the user's information needs. Instead, systems for \emph{query refinement} -- also known as \emph{query recommendation} or \emph{query suggestion} \cite{Sordoni2015AHR,BaezaYates2004QueryRU} -- can assist users by offering a set of followup queries that clarify and focus the user's search, progressively drilling down on the topics and entities of most interest (Fig. \ref{fig:teaser}).

\input{fig/teaser_wrapper.tex}

With this motivation in mind, we propose the task of \emph{entity-centric query refinement}. The task input is a list-intent query. The task output is a collection of $k$ \emph{query refinements}, referred to as a \emph{refinement set}. The refinement set should provide a reasonably comprehensive overview of the set of entities answering the input query. For instance, Fig. \ref{fig:teaser_b} shows an example of $k=4$ refinements that could familiarize the system user with some common types of pretrained NLP models, and point the user in interesting new search directions.

To our knowledge, no datasets are available which provide instances of list-intent queries paired with refinement sets satisfying our task goals. Therefore, we propose a method to create query / refinement set pairs which can be used to train a model for this task. We leverage the YAGO3 \cite{Mahdisoltani2015YAGO3AK} knowledge base, using YAGO entity types as training queries. YAGO types are based on the Wikipedia category system, which provides a rich, crowdsourced taxonomy of real-world entity types. Given a YAGO type, we consider all subtypes in the taxonomy as potential refinements (Fig. \ref{fig:teaser_a} shows an example). We propose a method \textbf{Q}uery \textbf{R}efinement via \textbf{E}ntity \textbf{S}pace \textbf{P}artitioning (\sysname), which selects as refinements the $k$ subtypes which provide the most comprehensive and non-redundant summary of the entities answering the input query. In head-to-head comparisons, we find that human annotators prefer refinement sets chosen using our proposed method over refinement sets consisting of $k$ randomly-chosen subtypes of the input query.

We use the resulting dataset to train a T5 model \cite{Raffel2020ExploringTL} capable of generating a refinement set for any input query. Since no evaluation dataset is available, we perform comparisons on both in-domain queries (held-out categories from the YAGO taxonomy) as well as out-of domain queries selected from Natural Questions \cite{Kwiatkowski2019NaturalQA} and the TREC 2009 Million Query Track \cite{Carterette2009Million}. We find that the outputs of a model trained on \sysname refinement sets are preferred over the outputs of a model trained on random query subtypes, suggesting that the properties captured by \sysname are generalizable to new queries. However, we also find that our models sometimes predict off-topic or irrelevant refinements, particularly on queries from Natural Questions and TREC. This points toward the need for future work to improve the reliability of refinement systems under domain shift, and to develop automated metrics of refinement quality which can be used to speed up the model development process.

In summary, our contributions are threefold: (1) We introduce the task of entity-centric query refinement, and outline key desiderata and evaluation criteria for this task. (2) We propose \sysname, which optimizes an entity-centric cost function to select refinement sets for queries covered by an existing knowledge base. The resulting refinement sets can be used both for entity exploration within the knowledge base, and as instances to train a refinement set generation model. (3) We develop a baseline model trained on instances selected by \sysname to generate refinement sets for queries unseen during training, and identify important areas for future work on this task based on analysis of our system outputs.

%% file: fig/teaser_wrapper.tex

\begin{figure}[t]
  \footnotesize
  \centering

  \begin{subfigure}[t]{0.39\columnwidth}
    \centering
    \includegraphics[width=\columnwidth]{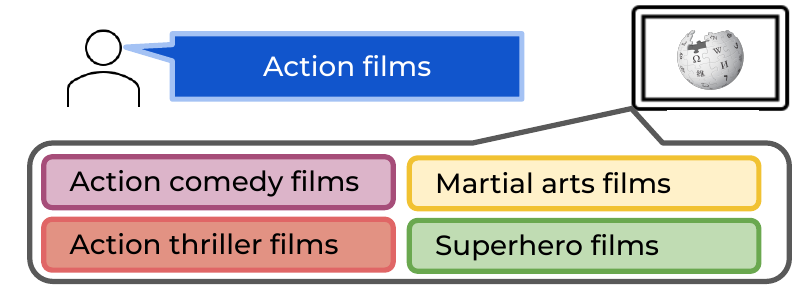}
    \caption{
      A query paired with a refinement set consisting of $k=4$ query subtypes available in an existing knowledge base.
    }
    \label{fig:teaser_a}
  \end{subfigure}
  \hspace{0.02\columnwidth}
  \begin{subfigure}[t]{0.51\columnwidth}
    \centering
    \includegraphics[width=\columnwidth]{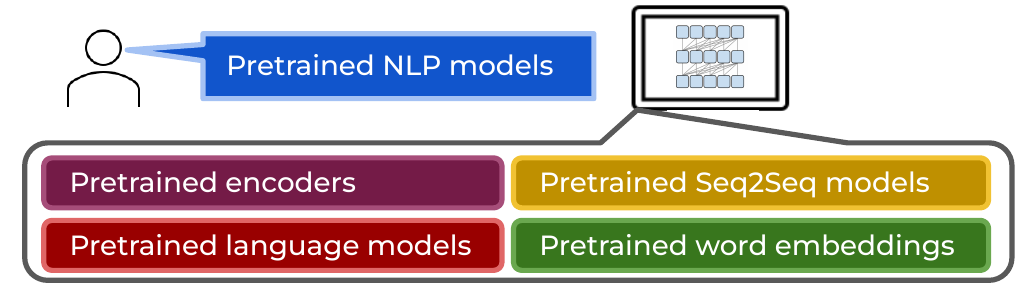}
    \caption{
      Refinements for a query unlikely to be covered by existing taxonomies, generated by a text-to-text model.
    }
    \label{fig:teaser_b}
  \end{subfigure}

  \vspace{-0.5em}

  \caption{
    Examples of the entity-centric query refinement task. We propose a method to select high-quality refinement sets for queries covered by an existing taxonomy (Fig. \ref{fig:teaser_a}), and use these refinement sets to train a model which can generate refinements for queries unlikely to be covered by any taxonomy (Fig. \ref{fig:teaser_b}).}
  \label{fig:teaser}

  \vspace{-0.1em}
\end{figure}

%% file: sec_task_definition.tex
\section{Task definition} \label{sec:task_definition}

\input{fig/entity_partition_wrapper.tex}

\subsection{Entity-centric query refinement} 

We aim to generate refinements which facilitate exploration and discovery for list-intent queries, helping the user drill down on entities of interest. Formally, the task input is a query $q$ whose answer is a list of entities. Equivalently, the query $q$ specifies an entity type. We refer to the list of entities answering $q$ as the \emph{answers} to $q$, or $\cA(q)$. The task output is a collection of $k$ \emph{refinements}\footnote{{In this work, we provide $k$ as a model input; dynamically choosing $k$ based on the query represents an important future research direction.}} $\cR(q) = \{ q_1', \dots, q_k' \}$. We refer to $\cR(q)$ as a \emph{refinement set}, or ``RS''. Each refinement $q_i'$ should itself be a list-intent query. In addition, each answer to $q_i'$ should be among the answers to $q$: $\cA(q_i') \subseteq \cA(q)$. In other words, each $q_i'$ should specify a subtype of $q$. For instance, every movie that is an answer to the refinement ``martial arts films'' is also an answer to the input query ``action films''.

\subsection{Desiderata for refinement sets} \label{sec:desiderata}

\input{table/example_evals_stage_1.tex}

The query refinement task is inherently open-ended, and there may be many reasonable RSs (refinement sets) for any given query. Inspired by prior work on faceted search interfaces for the Wikipedia category taxonomy \cite{Li2010FacetedpediaDG}, we conceptualize refinement generation from the standpoint of \emph{entity discovery}: given an input query $q$, can we design a refinement system such that any entity $e^* \in \cA(q)$ is discoverable after a few rounds of system interaction? From this standpoint, the best-possible refinement set would partition the entities $\cA(q)$ into $k$ disjoint, equally-sized subsets, such that each answer $e_j \in \cA(q)$ is an answer to exactly one refinement $q_i' \in \cR(q)$. This would ensure that any entity in $\cA(q)$ is discoverable after at most $\log_k(|\cA(q)|)$ refinements, since the refinements would induce a $k$-ary search tree over the entities answering $q$ (see Appendix \ref{appx:huffman}). We refer to such a refinement set as \emph{ideal}.

In practice, it will almost never be possible to generate an ideal RS, since each refinement must specify a semantic category expressed in natural language. Fig. \ref{fig:entity_partition} provides an example showing how the refinements for the query ``action films'' from Fig. \ref{fig:teaser} provide a good approximation to an ideal RS. We formalize this notion in \S \ref{sec:method}.

\subsection{Evaluation criteria} \label{sec:evaluation_criteria}

Efficient entity discovery provides motivation for our task formulation, but does not admit a simple evaluation strategy. Therefore, to assess the usefulness of proposed refinement sets, we conduct A / B tests where annotators compare two competing refinement sets $\cRa(q)$ and $\cRb(q)$ on a number of attributes. The evaluation includes two stages.

\paragraph{Stage 1: Validity of individual refinements} First, the annotator confirms that the individual refinements making up $\cRa(q)$ and $\cRb(q)$ conform to the task definition, requiring:

\begin{enumerate}[leftmargin=*,noitemsep]
    \item \textbf{Fluency}: Each refinement must be fluent and grammatical. 
    \item \textbf{Relevance}: Each refinement must specify a subtype of $q$. Non-fluent refinements are automatically judged as not relevant.
\end{enumerate}

Table \ref{tbl:evals_individual} provides some examples of queries that pass and fail these requirements. If fewer than half of the refinements from either RS satisfy the criteria, annotation stops here. In our experiments (\S \ref{sec:generation}), we find that virtually all refinements are fluent, and the Stage 1 screen serves in practice to filter out irrelevant refinements.

\paragraph{Stage 2: Overall refinement set quality} If the majority of refinements in both $\cRa(q)$ and $\cRb(q)$ are judged valid, the annotator compares the two RSs, based on four attributes:

\begin{enumerate}[leftmargin=*,noitemsep]
    \item \textbf{Comprehensiveness}: Does $\cR(q)$  provide a good overview of the entities answering the query $q$?
    \item \textbf{Interestingness}: Do the refinements in $\cR(q)$ provide new information about the different kinds of entities answering $q$, or are they generic and uninteresting?
    \item \textbf{Non-redundancy}: Does each refinement in $\cR(q)$ specify a unique entity type, or are some of them redundant?
    \item \textbf{Overall usefulness}: Overall, how useful are the refinements $\cR(q)$ for learning more about the entities answering $q$?
\end{enumerate}

Table \ref{tbl:evals_group} provides examples. For each attribute, the annotator selects whether $\cRa$ is better, $\cRb$ is better, or whether the two RSs are equally good. Details of the annotation process are provided in \S \ref{sec:human_eval_gold}. The full annotation guide is included in Appendix \ref{appx:annotation_guide}.

\input{table/example_evals_stage_2.tex}

%% file: fig/entity_partition_wrapper.tex

\begin{wrapfigure}{r}{0.40\textwidth}
\centering
\vspace{-5em}
\includegraphics[width=0.38\textwidth]{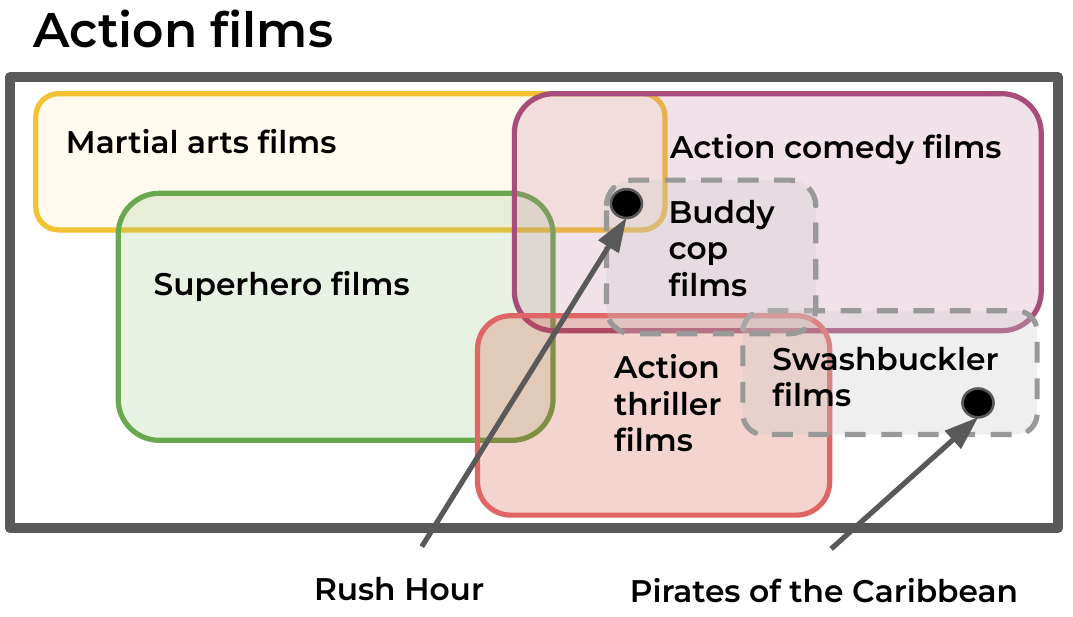}
\caption{
  An entity-space view of a refinement set for the query ``Action films''. Rectangles indicate refinements, and black circles indicate entities. The four rectangles with solid borders correspond to the refinements in Fig. \ref{fig:teaser}. The rectangles with dashed borders show two subgenres that were not included in the refinement set, since they are redundant / do not cover many films. The four selected refinements approximately partition the answer space.
}
\label{fig:entity_partition}
\vspace{-2em}

\end{wrapfigure}

%% file: table/example_evals_stage_1.tex
\begin{wraptable}{r}{0.42\textwidth}
    \footnotesize
    \vspace{-2em}
    \begin{tabular}{
            L{10em}
            R{3em}
            R{3em}
        }
        \toprule
        \multicolumn{3}{l}{\textbf{Query}: Action films} \\
        \midrule
        Refinement          & Fluent & Relevant      \\
        \midrule
        Martial arts films      & \cmark & \cmark        \\
        Romance films           & \cmark & \xmark        \\
        Martially flims arts of & \xmark & \xmark        \\
        \bottomrule
    \end{tabular}
    \caption{\textbf{Stage 1 evaluation} screens out individual refinements which are not fluent and relevant.}
    \label{tbl:evals_individual}
    \vspace{-3em}
\end{wraptable}

%% file: table/example_evals_stage_2.tex
\begin{table}[t]
    \footnotesize
        \begin{tabular}{
                L{0.1em}
                L{28em}
                R{4em}
                R{4em}
                R{4em}
            }
            \toprule
            \multicolumn{5}{l}{\textbf{Query}: Action films}                                                                      \\
            \midrule
              & Refinement set                                                & Compre-hensive & Inter-esting & Non-redundant \\
            \midrule
            1 & Action comedies, Action thrillers, Martial arts films, Spy films  & \cmark         & \cmark       & \cmark        \\
            \cmidrule(lr){1-5}
            2 & Action films set in \{Asia, North America, Africa, Europe\}       & \cmark         & \xmark       & \cmark        \\
            \cmidrule(lr){1-5}
            3 & Karate films, Kung Fu films, Boxing films, Films with boxing & \xmark         & \cmark       & \xmark        \\
            \bottomrule
        \end{tabular}
        \caption{\textbf{Stage 2 evaluation} assesses the overall quality of refinement sets. The notation ``\{Asia, North America, \dots\}'' means ``Action films set in Asia, Action films set in North America, \dots''. Row (2) is comprehensive since many action films take place on one of the listed continents, but is not interesting since many different kinds of queries can be categorized by continent. Row (3) is redundant and not comprehensive since it only covers martial arts movies, and repeats ``boxing films''. Human evaluation makes comparisons between two refinement sets, rather than binary \cmark / \xmark\ decisions for a single set; we show binary decisions for illustration.}
        \label{tbl:evals_group}
\end{table}

%% file: sec_method.tex
\section{Refinement set selection} \label{sec:method}

To build a baseline system for RS generation, we proceed as follows: (1) Leverage an existing knowledge base to create a collection of query / RS pairs that are close to ideal, and (2) Train a text-to-text model on this collection, with the goal of generalizing to queries not covered by a taxonomy. We describe (1) in this section and \S \ref{sec:dataset}, and describe (2) in \S \ref{sec:generation}.

\paragraph{Source taxonomy}

We use the YAGO3 taxonomy \cite{Suchanek2007YagoAC,Mahdisoltani2015YAGO3AK} (referred to simply as YAGO) as our source to construct a training dataset. The YAGO entity type system is adopted from the Wikipedia category system, a crowdsourced polyhierarchy used to organize Wikipedia pages. We use types in the YAGO taxonomy as input queries $q$. Given a type $q$, we consider all sub-types of $q$ in the taxonomy as refinement \emph{candidates}, denoted $\cC(q)$. From this collection of $K$ candidates, we aim to select $k$ sub-types to form our refinement set $\cR(q)$. We use the entities in the YAGO knowledge base that are instances of type $q$ as the answers to $q$, or $\cA(q)$. As a concrete example, in Fig. \ref{fig:entity_partition}, ``Action films'' is the query $q$, the shaded rectangles correspond to members of the candidate set $\cC(q)$, and the black points are two answers in $\cA(q)$.

Even when a list of $K$ candidate subtypes is available, selecting the best $k$ refinements is challenging because (1) $K$ may be large; for example, there are 68 subtypes of the type ``Action films'', (2) Many of the candidate subtypes may be too specific to be part of a useful summary (e.g. ``Tomb Raider films'' is a subtype of ``Action films''; Appendix \ref{appx:yago3_examples} provides additional examples), and (3) subtypes may be redundant, overlapping, or generic.

\paragraph{\sysname for refinement selection}

\setlength{\abovedisplayskip}{4pt}
\setlength{\belowdisplayskip}{4pt}

To select an RS of $k$ refinements from among the $K$ candidate subtypes associated with a given input query, we propose \sysname: \textbf{Q}uery \textbf{R}efinement via \textbf{E}ntity \textbf{S}pace \textbf{P}artitioning. We define a cost function which selects RSs according to the desiderata described in \S \ref{sec:desiderata}, rewarding RSs whose answers approximately partition the answers to the original query. Then, we minimize this cost function over all refinement sets. 

Let $e_j$ denote a particular entity in $\cA(q)$, and let $n = |\cA(q)|$. Given a pool of candidate refinements $\cC(q)$, let $\fR(q)$ indicate the collection of all size-$k$ subsets of $\cC(q)$. Given a possible refinement set $\cR(q) \in \fR(q)$, and a refinement $q'_i \in \cR(q)$, let $\aij = \ind \left[ e_j \in \cA(q'_i) \right]$. Define $c_j = \sumI(\aij)$, the number of refinements in $\cR(q)$ for which entity $j$ is an answer. Let $n_i = \sumJ(\aij) = |\cA(q'_i)|$, the number of entities that answer refinement $i$. Then we define a cost function $\cS$ measuring the quality of a given $\cR(q)$, and minimize over $\fR(q)$:
\vspace{-2em}
\begin{multicols}{2}
    \begin{equation}
        \cS(\cR(q)) = \left[ \sum_{j=1}^{n} | c_j - 1 | \right] - \min_{i \in \{1, \dots, k\}} n_i \label{eqn:cost}
    \end{equation}\break
    \begin{equation}
        \cR^*(q) = \argmin_{\fR(q)} \cS(\cR(q)). \label{eqn:best}
    \end{equation}
\end{multicols}

The first term in Eq. \ref{eqn:cost} is minimized when each answer to $q$ is an answer to exactly one of the $q'_i$, encouraging comprehensiveness and non-redundancy. The second term encourages the smallest refinement to have as many answers as possible. Combined with the first term, this rewards the selection of refinements which all have a similar number of answers. 

Eq. \ref{eqn:best} selects the refinement set $R^*(q)$ minimizing Eq. \ref{eqn:cost}. In Appendix \ref{appx:proof}, we provide a proof that the RS selected by \sysname is the \emph{best achievable} in the following sense: the scoring function $\cS(\cR(q))$ achieves its global minimum if and only if $\cR(q)$ is \emph{ideal} as defined in \S \ref{sec:desiderata}. Thus, selecting refinements according to \sysname yields the RS that is closest to ideal, given the subcategories $\cC(q)$ available in YAGO. We use integer linear programming to perform the combinatorial optimization over $\fR(q)$ in Eq. \ref{eqn:best}; see Appendix \ref{appx:optim} for details.






%% file: sec_dataset.tex
\section{A dataset for entity-centric query refinement} \label{sec:dataset}

We apply \sysname to YAGO to create a dataset for entity-centric query refinement, and conduct A / B tests to assess whether our approach aligns with human judgments. We set the number of refinements to $k=5$ for all experiments; this is large enough to enable diverse refinement sets, but not so large as to be overwhelming for users.

\subsection{Dataset creation} \label{sec:dataset_creation}

\input{table/predictions/data_creation.tex}

We apply \sysname to select RSs from the YAGO taxonomy. We use YAGO types with at least 50 answer entities as our training queries, as it may be difficult to select non-trivial refinements for types with only a few answers. Given a type $q$ with candidate subtypes $\cC(q)$, we use rule-based filters to remove candidates that differ from $q$ only by the addition of a date, location, or a gender (e.g. ``singers'' $\rightarrow$ ``female singers''), as these tend to lead to generic refinements (see Appendix \ref{appx:rules} for the full list of rules). We refer to the filtered collection as $\cCFilter(q)$. Queries with fewer than $k$ subtypes post-filtering are removed. For the remaining queries, we apply \sysname to the candidates $\cCFilter(q)$ to select refinements $\Rsys(q)$. We call the resulting dataset $\Dsys = (q_i, \Rsys(q_i))_{i=1}^N$. We also construct two ablation datasets for comparison. For $\Drandom$, we select refinements by randomly choosing k subtypes from $\cC(q)$, without filtering. For $\Dfilter$, we choose $k$ random subtypes from $\cCFilter(q)$. Table \ref{tbl:example_data_creation} and Appendix \ref{appx:yago3_examples} provide examples. In Appendix \ref{appx:qresp_scores}, we present results confirming that RSs selected using \sysname consistently achieve lower costs as measured by Eq. \ref{eqn:cost}, compared to randomly-chosen RSs. Our training dataset consists of $\numQueriesFiltered$ query / RS instances for $\Dsys$ and $\Dfilter$, and $\numQueriesUnfiltered$ instances for $\Drandom$. 

We hold out \numQueriesDev query / RS instances to be used for model development in \S \ref{sec:generation}. Development queries and their subtypes are excluded from the train set. We select dev set queries by randomly sampling YAGO types with at least \minrefinementsDev subtypes, each of which must have at least \minAnswersDev answer entities. We use categories with many subtypes and answers because these are the categories for which query refinement has the most potential to facilitate user exploration and discovery. From our development set, we manually select a collection of \numQueriesYagoEval examples to be used for human evaluation in A / B tests. We select a diverse set of interesting, open-ended queries across a variety of domains, intended to mirror the types of queries likely to be seen in real-world use. Appendix \ref{appx:dataset} includes a discussion of all data selection and preprocessing choices, as well as additional statistics on YAGO. Appendix \ref{appx:eval_queries} includes the full list of queries used for human evaluation.

\subsection{Human evaluation} \label{sec:human_eval_gold} 

We compare refinements from  $\Rsys$ with refinements chosen randomly, $\Rrandom$. In addition, we compare $\Rsys$ to $\Rfilter$, to confirm that the improved ratings are due to selection using \sysname, and not simply the filtering out of uninteresting candidates. Details on the annotation process are included in Appendix \ref{appx:annotation}. 

A / B test results are shown in Table \ref{tbl:a_b_gold}. We perform statistical significance tests for all human comparisons; see Appendix \ref{appx:stats} for details on the statistical procedures used. More than 97\% of RSs from all approaches pass the Stage 1 screen for relevance and fluency. This is expected, since the refinements for training queries come from the YAGO taxonomy. In Stage 2, RSs selected by \sysname are preferred by annotators 86\% of the time over random RSs, and 73\% of the time over random RSs post-filtering. $\Rsys$ is preferred more frequently for comprehensiveness than for interestingness, likely because Eq. \ref{eqn:cost} explicitly rewards comprehensiveness. Overall, the results indicate that \sysname captures human intuitions about refinement quality.  

\input{table/a_b_gold_wrapper.tex}

%% file: table/predictions/data_creation.tex
\begin{table*}[t]
    \footnotesize
    \centering

    \begin{tabular}{
            L{4em} |
            R{40em}
        }
        \toprule
        \multicolumn{2}{c}{\textbf{Query}: Action films}                                                                          \\
        \midrule
        $\Dsys$    & \{Action comedy, Action thriller, Martial arts, Science fiction action, Spy\} films                        \\
        \cmidrule(lr){1-2}
        $\Drandom$ & \{1910s, 1920s, Australian, Brazilian, Nigerian\} action films                                             \\
        \cmidrule(lr){1-2}
        $\Dfilter$ & \{Action comedy, Action thriller, Animated action, The purge\} films, Action films based on actual events. \\
        \bottomrule
    \end{tabular}
    \caption{
        Refinement sets from $\Dsys$, $\Drandom$, and $\Dfilter$. $\Drandom$ includes many refinements based on a time period or a country of origin, which does not provide interesting new information about the topic. $\Dfilter$ is overly specific (e.g. ``The Purge films''), and thus does not provide as good an overview as $\Dsys$.
    }
    \label{tbl:example_data_creation}
\end{table*}

%% file: table/a_b_gold_wrapper.tex
\begin{table}[t]
    \footnotesize
    \centering





    \input{table/ab-gold-combined.tex}

    \caption{
        A / B tests comparing refinement sets from $\Dsys$ against $\Drandom$ (left) and $\Dfilter$ (right). $N$ indicates the number of annotated instances. For Stage 1 evaluation, we report the percentage of refinement sets from each system that passed the Stage 1 screen. For Stage 2 evaluation, we report the percentage of queries for which annotators preferred refinements from System A vs. System B. The ``non-redundant'' evaluation was added after these tests were performed, and is left blank. $^{**}$ indicates significance at $p < 0.001$.}
        \label{tbl:a_b_gold}
\end{table}

%% file: table/ab-gold-combined.tex
\begin{tabular}{
        L{3.8em}
        L{8.4em}
            *{3}{L{3.5em}}
        L{0.2em}
        *{3}{L{3.5em}}
    }
    \toprule
                                &                   & \multicolumn{3}{c}{\textbf{A} = \sysname vs. \textbf{B} = Random} &  & \multicolumn{3}{c}{\textbf{A} = \sysname vs. \textbf{B} = \filtername} \\
    \cmidrule(lr){3-5}  \cmidrule(lr){7-9}
                                & $N=105$           & \textbf{A}           & Neutral & \textbf{B}                       &  & \textbf{A}           & Neutral & \textbf{B}                            \\
    \midrule
    \textbf{Stage 1}            & Fluent + Relevant & \textbf{98}\%        & -       & 97\%                             &  & 100\%                & -       & 100\%                                 \\
    \cmidrule(lr){1-9}
    \fourrows{\textbf{Stage 2}} & Comprehensive     & \textbf{88\%}$^{**}$ & 10\%    & 2\%                              &  & \textbf{75\%}$^{**}$ & 15\%    & 10\%                                  \\
                                & Interesting       & \textbf{71\%}$^{**}$ & 26\%    & 3\%                              &  & \textbf{64\%}$^{**}$ & 24\%    & 12\%                                  \\
                                & Non-redundant     & -                    & -       & -                                &  & -                    & -       & -                                     \\
                                & Overall           & \textbf{86\%}$^{**}$ & 10\%    & 4\%                              &  & \textbf{73\%}$^{**}$ & 17\%    & 10\%                                  \\
    \bottomrule
\end{tabular}

%% file: sec_generation.tex
\vspace{-0.5em}

\section{Refinement generation} \label{sec:generation}

Having established that \sysname selects high-quality refinement sets for queries covered by a knowledge base, we finetune a pretrained language model to generate refinements for queries not found in the KB. We experiment with two evaluation datasets: held-out categories from the YAGO taxonomy, and a collection of list-intent queries selected from Natural Questions \cite{Kwiatkowski2019NaturalQA} and the TREC 2009 Million Query Track \cite{Carterette2009Million}. 

\vspace{-0.5em}

\subsection{Model}

We use T5-3B \cite{Raffel2020ExploringTL} as our base model. Given a dataset $\cD$ consisting of training pairs $(q_i, \cR(q_i))_{i=1}^N$, we provide $q$ as the input for T5 and train it to generate $\cR(q)$. We format $\cR(q)$ by concatenating the individual refinements in alphabetical order, separated by a sentinel token. At prediction time, we provide $q$ as input and greedily decode (greedy outperformed sampling and beam search on automated metrics). We train models on $\Dsys$, $\Dfilter$, and $\Drandom$; we call these $\Msys$ , $\Mfilter$, and $\Mrandom$, respectively. We train with a batch size of 32 for 8000 steps, using the default T5 optimizer. 

As an additional ablation, we train a model $\Mseparate$ on $\Dsys$, which predicts a single refinement $q_i'$ at a time, rather than predicting a full refinement set $\cR(q)$. At prediction time, given an input $q$, we sample 5 separate predictions from $\Mseparate$ and concatenate to form $\cR(q)$; a similar approach is used by \citet{MacAvaney2021IntenT5SR} to generate possible query intents for  search result diversification. 

\vspace{-0.5em}

\subsection{Automated evaluations} \label{sec:auto_eval}

\input{table/auto_evals/automated_metrics.tex}

The results of \S \ref{sec:dataset} indicate that human annotators prefer refinements from $\Dsys$ over those from $\Drandom$ and $\Dfilter$. Therefore, for our automated evaluations, we treat RSs from the $\Dsys$ dev set as ``silver'' data against which to evaluate the predictions of our trained models. We evaluate using the following metrics:

\begin{enumerate}[leftmargin=*,noitemsep]
    \item \textbf{Sequence-based metrics}: We treat a generated RS as a text sequence with no additional structure, and compare it to the corresponding silver RS using BLEU and ROUGE-L (ROUGE-1 and -2 showed the same trend).
    \item \textbf{Set-based metrics}: We treat the generated RS as a set of $k$ refinements, and evaluate the precision, recall, and F1 relative to the set of silver refinements.
    \item \textbf{Perplexity}: We evaluate the perplexity of the silver RSs, formatted as a sequence, under each generation model.
\end{enumerate}

The results are shown in Table \ref{tbl:auto_eval}. $\Msys$ performs best on all metrics. Training on randomly-chosen candidate subtypes ($\Mfilter$ and $\Mrandom$) decreases performance, particularly as measured by F1. We also observe that sequential refinement set generation by $\Msys$ outperforms separate prediction of individual refinements by $\Mseparate$. We examine the reasons for this improvement in \S \ref{sec:eval_human}.

\subsection{Human evaluation} \label{sec:eval_human}

\input{table/predictions/yago_pred.tex}
\input{table/ab_yago_pred_wrapper.tex}

We conduct A / B tests on two datasets. First, we evaluate on the (in-domain) human evaluation set of YAGO types described in \S \ref{sec:dataset_creation}. Second, to measure the ability to generalize to \emph{out-of-domain} queries, we evaluate on \numQueriesNQEval real-world list-intent queries selected by one of the paper authors from the Natural Questions dataset and the TREC 2009 Million Query Track (referred to as \nqtrec). As with YAGO, we aimed to select exploratory, list-intent queries on a variety of topics. Appendix \ref{appx:eval_queries} includes the full list of \nqtrec queries, and Appendix \ref{appx:nqtrec_selection} includes a discussion of our query selection criteria.

\paragraph{Results on YAGO}

\input{table/ab-nq-best-random_wrapper.tex}
\input{table/predictions/nq_pred.tex}

We compare refinements from $\Msys$ against $\Mrandom$, $\Mfilter$ and $\Mseparate$. Table \ref{tbl:yago_pred_example} shows the predictions of each system on a single query. Table \ref{tbl:ab_yago_pred} shows the results of A / B tests comparing $\Msys$ against $\Mrandom$ and $\Mseparate$; results for $\Mfilter$ are similar to $\Mrandom$ and are included in Appendix \ref{appx:human_eval}. 

Compared to $\Mrandom$, refinements from $\Msys$ are much more likely to be comprehensive and interesting. On the other hand, the models are similar in terms of non-redundancy, since $\Mrandom$ tends to offer overly-specific refinements which provide a poor summary but do not overlap. $\Msys$ and $\Mseparate$ have nearly identical interestingness. However, $\Msys$ enjoys a large advantage in non-redundancy, since it can condition each new refinement on earlier refinements in the RS.

\paragraph{Results on \nqtrec} 

We compare predictions from $\Msys$ vs. $\Mrandom$. Interestingly, only 56\% of RSs generated by $\Msys$ pass the Stage 1 screen, compared to 81\% for $\Mrandom$. However, the RSs from $\Msys$ that pass the screen are preferred over RSs from $\Mrandom$ in terms of comprehensiveness, interestingness, and overall quality (Table \ref{tbl:ab_nq_best_random}). While the differences do not reach the threshold of statistical significance, this trend suggests that the two models behave differently on out-of-domain queries. $\Msys$ refinements can be interesting and informative (Table \ref{tbl:nq_pred_example_good}), but sometimes go off-topic (Table \ref{tbl:nq_pred_example_bad}). $\Mrandom$ tends to make ``safe'' refinements that are generic but relevant.

%% file: table/auto_evals/automated_metrics.tex
\begin{table}[t]
    \footnotesize
    \centering

    \begin{tabular}{
            L{4em}
            R{4.8em}
            R{4.8em}
            *{3}{R{1.8em}}
            R{4.5em}
        }
        \toprule
        \multirow{3}{*}{Model} & \multicolumn{2}{c}{Sequence}  & \multicolumn{3}{c}{Set}                       & \multirow{3}{*}{Perplexity} \\
        \cmidrule(lr){2-3}  \cmidrule(lr){4-6}
                               & BLEU          & ROUGE-L       & P             & R             & F1            &                             \\
        \midrule
        $\Msys$                & \textbf{67.1} & \textbf{69.4} & \textbf{24.8} & \textbf{24.3} & \textbf{24.6} & \textbf{2.01}               \\
        \cmidrule(lr){1-7}
        $\Mseparate$           & 66.9          & 68.2          & 19.6          & 19.2          & 19.4          & 2.35                        \\
        $\Mfilter$             & 61.5          & 66.0          & 15.1          & 14.3          & 14.7          & 2.11                        \\
        $\Mrandom$             & 57.6          & 63.8          & 6.8           & 6.7           & 6.8           & 2.19                        \\
        \bottomrule
    \end{tabular}
    \caption{Automated evaluation of generation models, using $\Dsys$ as ``silver'' evaluation targets. Evaluations are categorized into Sequence-based, Set-based, and Perplexity-based. $\Msys$ outperforms models trained on randomly-chosen refinements, or trained to generate refinements separately rather than as a single sequence.}
    \label{tbl:auto_eval}
\end{table}

%% file: table/predictions/yago_pred.tex
\begin{table*}[t]
    \footnotesize
    \centering

    \begin{tabular}{
            L{5em} |
            R{39em}
        }
        \toprule
        \multicolumn{2}{c}{\textbf{Query}: Physicians} \\
        \midrule
        $\Msys$ & Alternative medicine physicians, cardiologists, dermatologists, ophthalmologists, psychiatrists  \\
        \cmidrule(lr){1-2} 
        $\Mrandom$ & \{Canadian, German, Norwegian\}  dermatologists, Pediatricians, Women physicians \\
        \cmidrule(lr){1-2}
        $\Mfilter$ & Fictional physicians, Oncologists, Psychiatric physicians, Radiologists, surgeons
         \\
        \cmidrule(lr){1-2}
         $\Mseparate$ & \{Atheist, Baritone, Fictional, Military\} physicians, Neurologists \\
        \bottomrule
    \end{tabular}
    \caption{
        Refinements of $\Msys$ and three ablations on a YAGO evaluation query. The $\Msys$ suggestions cover 5 common types of physicians. In contrast, some ablation refinements are generic (Canadian dermatologists) or idiosyncratic (Fictional physicians).
    }
    \label{tbl:yago_pred_example}
\end{table*}

%% file: table/ab_yago_pred_wrapper.tex
\begin{table}[t]
    \footnotesize
    \centering

    \input{table/ab-yago-combined.tex}




    \caption{
        Results of A / B tests on the YAGO human evaluation set. $\Msys$ is preferred over both ablations. $^{*}$ and $^{**}$ indicate significance at $p < 0.05$ and $p < 0.001$, respectively.
    }
    \label{tbl:ab_yago_pred}
\end{table}

%% file: table/ab-yago-combined.tex
\begin{tabular}{
        L{3.8em}
        L{8.4em}
            *{3}{L{3.5em}}
        L{0.2em}
        *{3}{L{3.5em}}
    }
    \toprule
                                &                   & \multicolumn{3}{c}{\textbf{A} = \sysname vs. \textbf{B} = Random} &  & \multicolumn{3}{c}{\textbf{A} = \sysname vs. \textbf{B} = Separate} \\
    \cmidrule(lr){3-5}  \cmidrule(lr){7-9}
                                & $N=105$           & \textbf{A}           & Neutral & \textbf{B}                       &  & \textbf{A}           & Neutral & \textbf{B}                         \\
    \midrule
    \textbf{Stage 1}            & Fluent + Relevant & 93\%                 & -       & \textbf{100}\%$^{*}$             &  & 93\%                 & -       & \textbf{95}\%                      \\
    \cmidrule(lr){1-9}
    \fourrows{\textbf{Stage 2}} & Comprehensive     & \textbf{77\%}$^{**}$ & 12\%    & 11\%                             &  & \textbf{47\%}        & 20\%    & 33\%                               \\
                                & Interesting       & \textbf{70\%}$^{**}$ & 22\%    & 8\%                              &  & \textbf{28\%}        & 45\%    & 27\%                               \\
                                & Non-redundant     & \textbf{22\%}        & 66\%    & 12\%                             &  & \textbf{42\%}$^{**}$ & 45\%    & 14\%                               \\
                                & Overall           & \textbf{76\%}$^{**}$ & 17\%    & 7\%                              &  & \textbf{46\%}$^{*}$  & 26\%    & 28\%                               \\
    \bottomrule
\end{tabular}

%% file: table/ab-nq-best-random_wrapper.tex
\begin{wraptable}{R}{0.47\textwidth}
    \footnotesize
    \centering

    \input{table/human_evals/ab-nq-filter_joint_best_gold_false-all_joint_random_gold_false.tex}
    \caption{Human evaluations on \nqtrec. $\Msys$ refinements tend to be more interesting and comprehensive, provided that they are relevant. $^{**}$ indicates $p < 0.001$.}
    \label{tbl:ab_nq_best_random}
\end{wraptable}

%% file: table/human_evals/ab-nq-filter_joint_best_gold_false-all_joint_random_gold_false.tex
\begin{tabular}{
    L{6.7em} *{3}{L{3.4em}}
  }
  \toprule
                                 & \multicolumn{3}{c}{\textbf{A} = \sysname vs. \textbf{B} = Random} \\
  \cmidrule(lr){2-4}
  $N = 93$                       & \textbf{A}     & Neutral     & \textbf{B}                         \\
  \midrule
  \tworows{Fluent + \\ Relevant} & \tworows{56\%} & \tworows{-} & \tworows{\textbf{81\%}$^{**}$}     \\
  \\
  \cmidrule(lr){1-4}
  Comprehensive                  & \textbf{45\%}  & 30\%        & 26\%                               \\
  Interesting                    & \textbf{40\%}  & 34\%        & 26\%                               \\
  Non-redundant                  & 23\%           & 51\%        & \textbf{26\%}                      \\
  Overall                        & \textbf{43\%}  & 28\%        & 30\%                               \\
  \bottomrule
\end{tabular}

%% file: table/predictions/nq_pred.tex
\begin{table*}[t]
    \footnotesize
    \centering

    \begin{subtable}[h]{\linewidth}

        \begin{tabular}{
                L{4em} |
                R{40em}
            }
            \toprule
            \multicolumn{2}{c}{\textbf{Query}: Functions of government}                                                                                                                                                                                  \\
            \midrule
            $\Msys$    & \{Agricultural, Defense, Education, Finance, Foreign\} functions of the government \\
            \cmidrule(lr){1-2}
            $\Mrandom$ & Functions of the \{Canadian, Yukon, Quebec, Northern Mariana Islands, United States\} government \\
            \bottomrule
        \end{tabular}
        \caption{$\Msys$ offers a list of five interesting, diverse government functions. $\Mrandom$ provides a generic list of five locations, including two Canadian provinces.}
        \label{tbl:nq_pred_example_good}

    \end{subtable}

    \begin{subtable}[h]{\linewidth}
        \begin{tabular}{
                L{4em} |
                R{40em}
            }
            \toprule
            \multicolumn{2}{c}{\textbf{Query}: Popular YouTube Channels}                                                                                                                                 \\
            \midrule
            $\Msys$    &
            \{Celebrity YouTube, Music video, Religious YouTube, Religious television, Religious video\} channels \\
            \cmidrule(lr){1-2}
            $\Mrandom$ & \{American, Australian, British, Japanese, Pakistani\} popular YouTube channels \\
            \bottomrule
        \end{tabular}

        \caption{Three $\Msys$ refinements were judged irrelevant since they don't mention YouTube specifically. $\Mrandom$ makes five generic, but relevant, refinements by once against listing five locations.}
        \label{tbl:nq_pred_example_bad}

    \end{subtable}
    
    \vspace{-0.5em}

    \caption{Predictions on two queries from NQ+TREC. $\Msys$ provides high-quality refinements for the first query, but is slightly off-topic for the second one.}
    \label{tbl:nq_pred_example}

\end{table*}

%% file: sec_related_work.tex
\vspace{-0.5em}

\section{Related work} \label{sec:related_work}

\vspace{-0.5em}

\paragraph{Query refinement}
 \emph{Search results clustering} \cite{Carpineto2009ASO} offers refinements by retrieving a collection of documents in response to an input query, clustering them, and assigning an informative name to each cluster. Names can be assigned based on word count statistics \cite{Zamir1999GrouperAD,Osinski2005ACA}, or generated using Seq2Seq models \cite{Medlar2021QuerySA}. \emph{Faceted search} \cite{Tunkelang2009FacetedS,Hearst2006ClusteringVF} also retrieves documents in response to the user's search, but organizes them according to a faceted concept hierarchy, which can be used as a source of refinements. The hierarchy can be created by the system designers \cite{Yee2003FacetedMF}, adapted from an existing taxonomy \cite{Li2010FacetedpediaDG,Arenas2016FacetedSO}, or constructed automatically \cite{Stoica2007AutomatingCO}. 

Like our work, these approaches output a collection of query refinements, aided by an existing taxonomy in the case of faceted search. Unlike these approaches, we optimize an entity-centric objective function to select a set of refinements which provides a comprehensive summary of the input query, rather than organizing a collection of retrieved documents. In addition, unlike faceted search, our proposed baseline can provide refinements for an arbitrary input query which may not be covered by an existing taxonomy. 

Systems have also been trained on \emph{search query logs} to predict likely followup queries given a user search history \cite{Sordoni2015AHR,Dehghani2017LearningTA,Boldi2008TheQG}. Our modeling goals differs from this setting in that (1) we do not assume access to query logs, which are often proprietary, and (2) we aim to output a set of refinements with a specific entity structure, rather than reproducing the search behavior of users.

\paragraph{Under-specified queries}
Researchers in NLP and IR have developed systems to resolve ambiguity \cite{Min2020AmbigQAAA}, incorporate dialogue context \cite{Elgohary2019CanYU,Anantha2021OpenDomainQA}, and ask questions \cite{Aliannejadi2019AskingCQ,Sekulic2021TowardsFG,Zamani2020GeneratingCQ} to clarify user intent in response to ambiguous or multi-faceted input queries. As an alternative to clarifying user intent directly, \emph{search results diversification} \cite{Santos2015SearchRD} predicts possible user intents with the aid of taxonomies \cite{Agrawal2009DiversifyingSR}, search logs \cite{Santos2010ExploitingQR}, or recently transformer language models \cite{MacAvaney2021IntenT5SR}, and then selects a collection of documents which comprehensively and non-redundantly addresses all predicted user intents \cite{Carbonell1998TheUO}. 

In general, these approaches were developed to distinguish between a handful of possible query intents, while our goal is to facilitate exploration and entity discovery for open-ended list-intent queries. Like search results diversification, we also aim for comprehensiveness and non-redundancy. However, we measure these quantities directly over sets of entities, rather than over collections of documents, whose similarity to each other and to the input query must be approximated using a metric like cosine distance.

%% file: sec_conclusion.tex
\vspace{-0.5em}

\section{Conclusion and future work}

In this work, we proposed the task of entity-centric query refinement. We developed \sysname to select high-quality refinement sets which partition the set of entities answering the input query, and showed that our methodology has good agreement with human judgments. We then demonstrated that a text-to-text model trained on \sysname-selected data was able to generate refinement sets for queries not found in an existing knowledge base. 

Our findings point toward two key open challenges for entity-centric query refinement. First, in \S \ref{sec:eval_human}, we found that $\Msys$ can sometimes generate off-topic or irrelevant refinements under domain shift. This points to a need for domain adaptation techniques which do not require supervised query / refinement set pairs. A second challenge is the development of high-quality automated evaluation metrics to speed the model development process. While we experimented with a number of evaluation metrics and found that they correlated with human judgments of refinement quality (\S \ref{sec:auto_eval}), the metrics we examined perform comparisons to a single reference refinement set. Given the open-endedness of the task, these approaches may not adequately reward refinements which a human would judge as high-quality, but which do not closely match the reference.

The $\sysname$ framework has the potential to facilitate progress on both challenges. In this work, we used the \sysname scoring function (Eq. \ref{eqn:cost}) to select refinement sets from a pool of candidates for which gold answer entities could be found in a knowledge base. In the future, this same scoring function could be used to evaluate refinement sets proposed by a generation model, using an entity-centric QA model \cite{Ling2020LearningCE,Fvry2020EntitiesAE} to predict the list of entities answering each refinement. This \sysname-QA score would serve as a flexible automated metric to compare the performance of refinement generation models. 

Further, the availability of a flexible automated metric for refinement set quality could be leveraged to improve out-of-domain generalization. For instance, the \sysname-QA score could be used as a reward signal to be optimized via reinforcement learning; this approach could steer the model away from generating irrelevant refinements. Similarly, \sysname-QA could be used at inference time to re-rank a list of generated candidate refinements, filtering out irrelevant or off-topic candidates. 

In summary, we believe that the entity-centric query refinement task presents a number of exciting research avenues for researchers. We hope that our dataset, modeling baselines, and analysis will motivate future work on this challenging and relevant task.

%% file: sec_acknowledgments.tex
\acks{Thanks to Livio Baldini Soares for help with the RELIC code, and to Andrew McCallum, Ming-Wei Chang, Rob Logan, Sam Oates, Taylor Curtis, and the Google Research Language and Google Search teams for helpful discussions and feedback. Thanks also to Donald Metzler and John Blitzer for comments on a draft of this work.}

%% file: appx_qresp_tacl.tex
\section{Additional details on \sysname} \label{appx:qresp}

\subsection{\sysname and entity discovery} \label{appx:huffman}

In \S \ref{sec:desiderata} we mentioned that, when interacting with a (hypothetical) system which outputs $k$ refinements whose answers evenly partition the answers to the input query $q$, any entity answering $q$ is discoverable after at most $\log_k(|\cA(q)|)$ rounds of system interaction. To see why, suppose we aim to discover a target entity $e^* \in \cA(q)$. When given $q$ as input, this ideal system would output $k$ refinements, each with $|\cA(q)| / k$ answers, and exactly one of which has $e^*$ as an answer. We would then input this refinement $q'$ as our new query, and the system would output $k$ new refinements, each with $|\cA(q)| / k^2$ answers, exactly one of which has $e^*$ as an answer. We would use this refinement $q''$ as our new input query, and repeat the process until we arrive at a refinement that includes only $e^*$. This process induces a $k-ary$ tree over the set of all entities answering the original query $q$, with depth $\log_k(|\cA(q)|)$. In other words, any entity answering $q$ can be discovered after $\log_k(|\cA(q)|)$ system interactions.

This type of system interaction can be viewed as a form of Huffman coding, where each entity $e_j \in \cA(q)$ is represented by a $k$-ary prefix code indicating the sequence of refinements used to discover the entity. Assuming that all entities in $\cA(q)$ are equally likely to be the target entity $e^*$, choosing refinements which partition the entity space into equal-sized subsets induces an optimal prefix code \cite[Chapter~16.3]{Cormen2009Introduction}. Future work could extend this framework to model entity popularity, assigning shorter codes (i.e. shorter refinement sequences) to more frequently-searched entities. 

\subsection{Best achievable refinements} \label{appx:proof}

We provide a proof that the scoring function $\cS(\cR(q))$ defined in Eq. \ref{eqn:cost} achieves its global minimum if and only if $\cR(q)$ is ideal -- i.e. all refinements in $\cR(q)$ have the same number of answers, and every entity answering $q$ answers exactly one $q'_i \in \cR(q)$.

For brevity, we denote an RS $\cR(q)$ simply as $\cR$. Assume $n \triangleq |\cA(q)|$ is divisible by $k$; this eliminates some edge cases but does not change the main idea. Re-write Eq. \ref{eqn:cost} by defining $t_1 \triangleq \sum_j |c_j - 1|$ and $t_2 \triangleq \min_i n_i$, so $\cS(\cR) = t1 - t2$. We claim that $\cR$ is ideal as defined in \S \ref{sec:desiderata} if and only if $t_1 = 0$ \emph{and} $t_2 = n / k$. As proof, the forward direction follows from the definition. For the reverse direction, $t_1 = 0$ means that each $e_j$ answers exactly one $q_i'$. Combined with the fact that the smallest $\cA(q'_i)$ has $n / k$ answers, this implies that \emph{all} $q_i'$ have $n / k$ answers, thus $\cR$ is ideal. It follows that $\cS(\cR) = - n / k$ if $\cR$ is ideal.

Now assume that $\cR$ is some RS for which $\cS(\cR) \leq -n / k$. Since $t_1$ is lower-bounded by 0, it must be that $t_2 \geq n / k$; equivalently, $t_2 = n / k + \delta$ for some integer $\delta \geq 0$. This necessitates that $|\cA(q'_i)| \geq n / k + \delta$ for all $i$, which implies $t_1 \geq \delta k$. Then $\cS(\cR) = t_1 + t_2 \geq \delta k - (n / k + \delta) = (k-1) \delta - n / k$. If $\delta > 0$, then $\cS(\cR) > -n / k$, contradicting our assumption. On the other hand, if $\delta = 0$, then we have $t_1 = 0$ and $t_2 = n / k$, which (from our earlier claim) is only possible if $\cR$ is ideal. Thus, $\cS(\cR)$ achieves its minimum if and only if $\cR$ is ideal.

\subsection{Optimization}  \label{appx:optim}

We describe how to convert the optimization problem Eq. \ref{eqn:best} into an integer linear program. We use the same notation as in \S \ref{sec:method}, with one change: instead of using $i$ to index refinements in $\cR(q)$, we use it to index refinement \emph{candidates} $q'_i \in \cC(q)$. Let $x_i = \ind \left[ q'_i \in \cR(q) \right]$, $\aij = \ind \left[ e_j \in \cA(q'_i) \right]$, and $c_j = \sum_i x_i \aij$. Let $n_i = |A(q'_i)|$ and $\nmax = \max_i |A(q'_i)|$. Then Eq. \ref{eqn:best} can be re-written as:

\begin{equation*} \label{eqn:ilp}
    \begin{aligned}
         & \min_{x_i, c_j, y_j, \xi} & \left[ \sum_{j=1}^{n} y_j \right] & - \xi                                                       \\
         & \textrm{s.t.}             & c_j - 1                           & \leq y_j \textrm{ and } 1 - c_j  \leq y_j \quad & \forall j \\
         &                           & c_j                               & = \sum_i x_i \aij \quad                         & \forall j \\
         &                           & k                                 & = \sum_i x_i                                                \\
         &                           & \xi                               & \leq (1 - x_i) \nmax + x_i n_i                  & \forall i
    \end{aligned}
\end{equation*}
We use SCIP to perform the optimization \cite{BestuzhevaEtal2021OO}.

%% file: appx_dataset.tex
\section{Dataset construction} \label{appx:dataset}

\subsection{YAGO source data}

The versions of the YAGO dataset used in this work can be downloaded from \url{https://www.mpi-inf.mpg.de/departments/databases-and-information-systems/research/yago-naga/yago/downloads}, under the heading \texttt{Download YAGO themes} and in the colored box labeled \texttt{TAXONOMY}. For the taxonomy (type hierarchy), we used \texttt{yagoTaxonomy}. For the list of entities that are instances of each type in the hierarchy, we used \texttt{yagoTransitiveType}. Entity type transitive closure is enforced; if $e_j$ is an instance of $q$, then it is also an instance of all ancestors of $q$. 

The full YAGO taxonomy is constructed by merging the WordNet taxonomy \cite{Miller1995WordNetAL} with the Wikipedia category hierarchy, using heuristics to label Wikipedia categories as subclasses of WordNet types. This merging process, while generally very effective, may be noisy or inaccurate in some cases. Since the Wikipedia category system already provides a rich, crowd-sourced taxonomy of real-world entity types, we remove the WordNet types for our experiments and focus on the Wikipedia category hierarchy.

\subsection{YAGO3 examples} \label{appx:yago3_examples}

Table \ref{tbl:yago_example} shows the YAGO category $q=$ ``Action films'', paired together with a sample of its sub-types $\cC(q)$ and answers $\cA(q)$. Many of the subtypes are generic or overly-specific and would not be suitable refinements. Table \ref{tbl:example_data_creation} in \S \ref{sec:dataset_creation} showed RSs from $\Dsys$, $\Drandom$, and $\Dfilter$ for a single query. In Table \ref{tbl:example_data_creation_extras}, we include examples for three additional queries.

\input{table/yago_example.tex}

\subsection{YAGO preprocessing steps} \label{appx:yago_preprocessing}

As described in \S \ref{sec:dataset_creation}, we perform a number of preprocessing steps before running \sysname to select RSs from YAGO. 

\begin{enumerate}[leftmargin=*,noitemsep]
    \item Use YAGO types with $\geq 50$ answer entities as training queries (discarding other types): This is done because the task we propose is intended to help users in situations where simply reading the answers to an input query would be overwhelming. For a query with a handful of answer entities, offering refinements is unlikely to be helpful.
    \item Filter out queries with fewer than $k=5$ subtypes: The majority of YAGO types have only a single subtype; Figure \ref{fig:category_distribution} shows the distribution of subtypes per parent type. Unfortunately, these queries cannot be sensibly used as training data, since our goal is to generate refinement sets including $k=5$  refinements. Fortunately, there is still enough data in the ``tail'' of this distribution to support a reasonably-sized training dataset.
    \item Filter out refinements which differ from the input query by the addition of generic modifiers (dates / locations / genders): This is done to focus on refinements which reveal some kind of interesting structure specific to the query in question. Given ``action films'' as an input, providing a list of action movie subgenres (e.g. ``martial arts films'', ``action comedy films'', etc.) reveals new information specific to the domain of the query, while ``action films from 1993'' or ``action films set in Great Britain'' does not. The full list of filtering rules is included in Appendix \ref{appx:rules}.
\end{enumerate}

The full YAGO dataset includes 187K Wikipedia types. Preprocessing leaves 8,958 queries for $\Dsys$ and $\Dfilter$, and 17,598 for $\Drandom$ ($\Drandom$ has more queries because there are some types with more than 5 subtypes in total, but fewer than 5 that pass pass the date / location / gender filters described above). The preprocessing steps are performed to ensure that the RSs in the final dataset are high-quality, suitable for model training, and represent realistic queries for which query refinement is a meaningful task.

\subsection{YAGO dev and evaluation set selection} \label{appx:yago_eval_data}

Our goal for our dev and test set is to select YAGO types that provide a realistic approximation of list-intent queries likely to be entered by users during search; these are also the types of queries where refinement selection approaches like \sysname have the potential to improve over a simpler approach. As such, for the dev set, we select YAGO types with a large number of subcategories (at least 15), each of which must have at least 200 answers. We require at least 15 candidate subtypes since \sysname (and methods like it) only has the potential to improve over random selection if there are a reasonable number of candidate categories to choose from. Similarly, we require 200 answers per candidate because trivial subtypes with only a handful of answers are clearly not suitable refinements (e.g. ``Pirates of the Caribbean films'' is not a good refinement for ``Action films''); the refinement selection task is much more challenging when it requires selecting among a large number of non-trivial potential refinements. 

For our human evaluation set, we manually selected queries from the dev set which we judged to be interesting and realistic. We also attempted to choose queries from a variety of subject areas, including sports, music, science, politics, entertainment, and technology; see Appendix \ref{appx:eval_queries} for the full list. 

\subsection{YAGO Filtering rules}  \label{appx:rules}

\input{table/predictions/data_creation_extras.tex}
\input{fig/category_distribution_wrapper.tex}

We use the following heuristics to obtain $\cCFilter(q)$ from $\cC(q)$.
For each $q'_i$ in $\cC(q)$, we compare $q$ to $q'_i$. We remove $q'_i$ from $\cCFilter(q)$ if it differed from $q$ by:

\begin{itemize}[leftmargin=*]
\item The addition of an entity tagged by Spacy as \texttt{DATE, GEP, NORP}, or \texttt{LOC}. (e.g. ``Politicians'' $\rightarrow$ ``American Politicians'').
\item The addition of a phrase matching one of the following regular expressions: 
\begin{itemize}[leftmargin=*]
    \item \texttt{[0-9]\{1,2\}(st|th)(-| )century}
    \item \texttt{1[0-9]\{3\}[\textasciicircum 0-9]}
    \item \texttt{[0-9]\{3\}[\textasciicircum 0-9]} 
\end{itemize}
    This filters out refinements like ``Politicians'' $\rightarrow$ $19^{th}$ century politicians.
\item The addition of one of the following: \texttt{male, female, men, women} (e.g. ``Politicians'' $\rightarrow$ ``Male politicians'').
\end{itemize}

\subsection{Checks on \sysname optimization} \label{appx:qresp_scores}

In Fig. \ref{fig:gold_scores_scatter_filter}, we plot the costs $\cS(\cR(q))$ of the RSs in $\Dsys$ versus those in $\Dfilter$. The results confirm that \sysname is able to identify RSs that have substantially lower cost than choosing randomly from $\cCFilter(s)$.

\input{fig/gold_scores_scatter_filter_wrapper.tex}

%% file: table/yago_example.tex
\begin{table}[t]
    \footnotesize
    \centering

    \begin{tabular}{
            L{8em} |
            R{38em}
        }
        \toprule
        Query $q$           & Action films                                                                                                                                                                             \\
        \cmidrule(lr){1-2}
        Candidates $\cC(q)$ & 1900s action films, 1910s action films, \dots, Action adventure films, Action comedy films, \dots, British action films, Chinese action films, \dots, The Purge films, Tomb Raider films \\
        \cmidrule(lr){1-2}
        Answers $\cA(q)$    & Bad Boys, John Wick, Rock Balboa, Foxy Brown, Rush Hour, Pirates of the Caribbean, Mad Max, \dots                                                                                        \\
        \bottomrule
    \end{tabular}
    \caption{An example of candidate refinements (i.e. sub-types) $\cC(q)$ and answers (i.e. instances) $\cA(q)$ for the example query ``Action films''. Some sub-types are specific to the domain (e.g. ``Action adventure films''), while others generically modify the query by adding a time period or country or origin.}
    \label{tbl:yago_example}
\end{table}

%% file: table/predictions/data_creation_extras.tex
\begin{table*}[t]
    \footnotesize
    \centering

    \begin{tabular}{
            L{4em} |
            R{44em}
        }
        \toprule
        \multicolumn{2}{c}{\textbf{Query}: Academic journals} \\
        \midrule
        $\Dsys$ & Healthcare journals, Humanities journals, John Wiley \& Sons academic journals, SAGE Publications academic journals, Scientific journals \\
        \cmidrule(lr){1-2}
        $\Drandom$ & Academic journals associated with non-profit organizations, Croatian-language journals, Latin-language journals, NRC Research Press academic journals, Ubiquity Press academic journals  \\
        \cmidrule(lr){1-2}
        $\Dfilter$ &  Berghahn Books academic journals, English-language journals, Multidisciplinary academic journals, Spanish-language journals, World Scientific academic journals \\
        \toprule

        \multicolumn{2}{c}{\textbf{Query}: Islands} \\
        \midrule
        $\Dsys$ & Disputed islands, New islands, River islands, Uninhabited islands, Volcanic islands \\
        \cmidrule(lr){1-2}
        $\Drandom$ &  Barrier islands, Islands of Europe, Islands of Sierra Leone, Mediterranean islands, Wikipedia categories named after islands \\
        \cmidrule(lr){1-2}
        $\Dfilter$ & Barrier islands, Coral islands, Former islands, Private islands, Volcanic islands \\
        \toprule

        \multicolumn{2}{c}{\textbf{Query}: Musicians} \\
        \midrule
        $\Dsys$ & Composers, Electronic musicians, Percussionists, Singers, Woodwind musicians \\
        \cmidrule(lr){1-2}
        $\Drandom$ & Australian musicians, Compost Records artists, Good Vibe Recordings artists, Inner Ear artists, Wax Trax! Records artists \\
        \cmidrule(lr){1-2}
        $\Dfilter$ & Black River Entertainment artists, DJM Records artists, Indianola Records artists, Pony Canyon artists, Ruthless Records artists \\
        \toprule

    \end{tabular}
    \caption{Additional example refinements from $\Dsys$, $\Drandom$, and $\Dfilter$.}
    \label{tbl:example_data_creation_extras}
\end{table*}

%% file: fig/category_distribution_wrapper.tex

\begin{wrapfigure}{R}{0.48\textwidth}
\centering
\footnotesize
\includegraphics[width=0.44\textwidth]{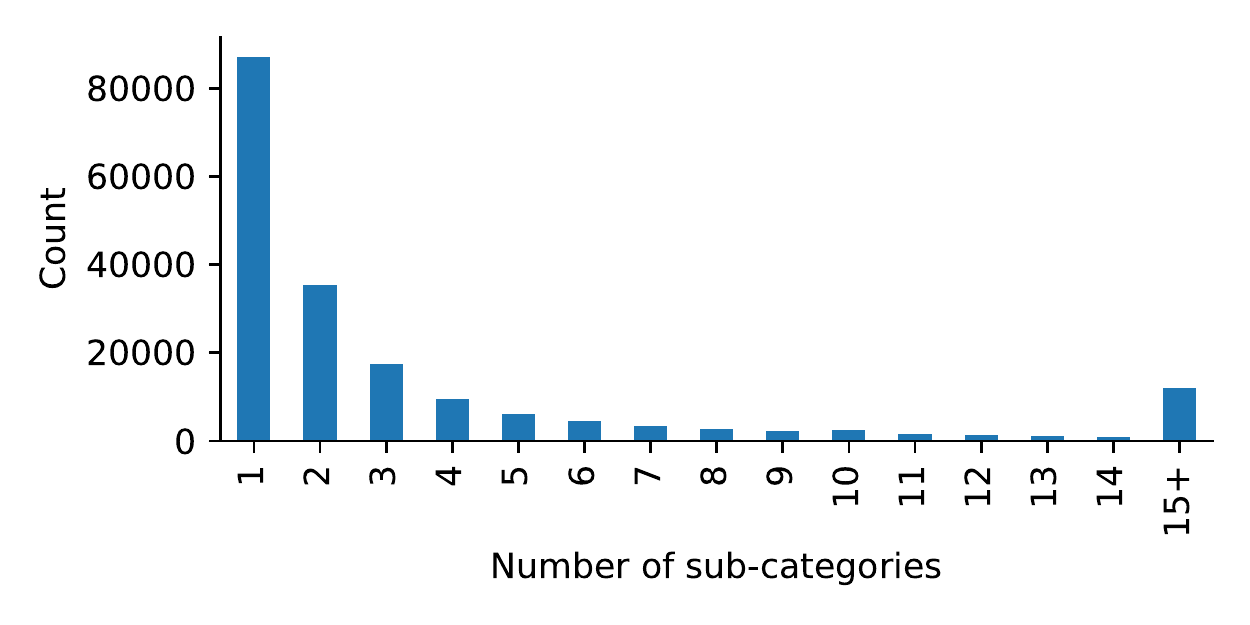}
\caption{
    YAGO3 category distribution. The x-axis indicates the number of sub-categories associated with a given YAGO3 category, and the heigh indicates the number of categories with that many subcategories. For instance, roughly 80K YAGO3 categories have exactly one subcategory. 
}
\label{fig:category_distribution}
\vspace{-2em}

\end{wrapfigure}

%% file: fig/gold_scores_scatter_filter_wrapper.tex
\begin{wrapfigure}{r}{0.42\textwidth}
  \centering
  \includegraphics[width=0.38\columnwidth]{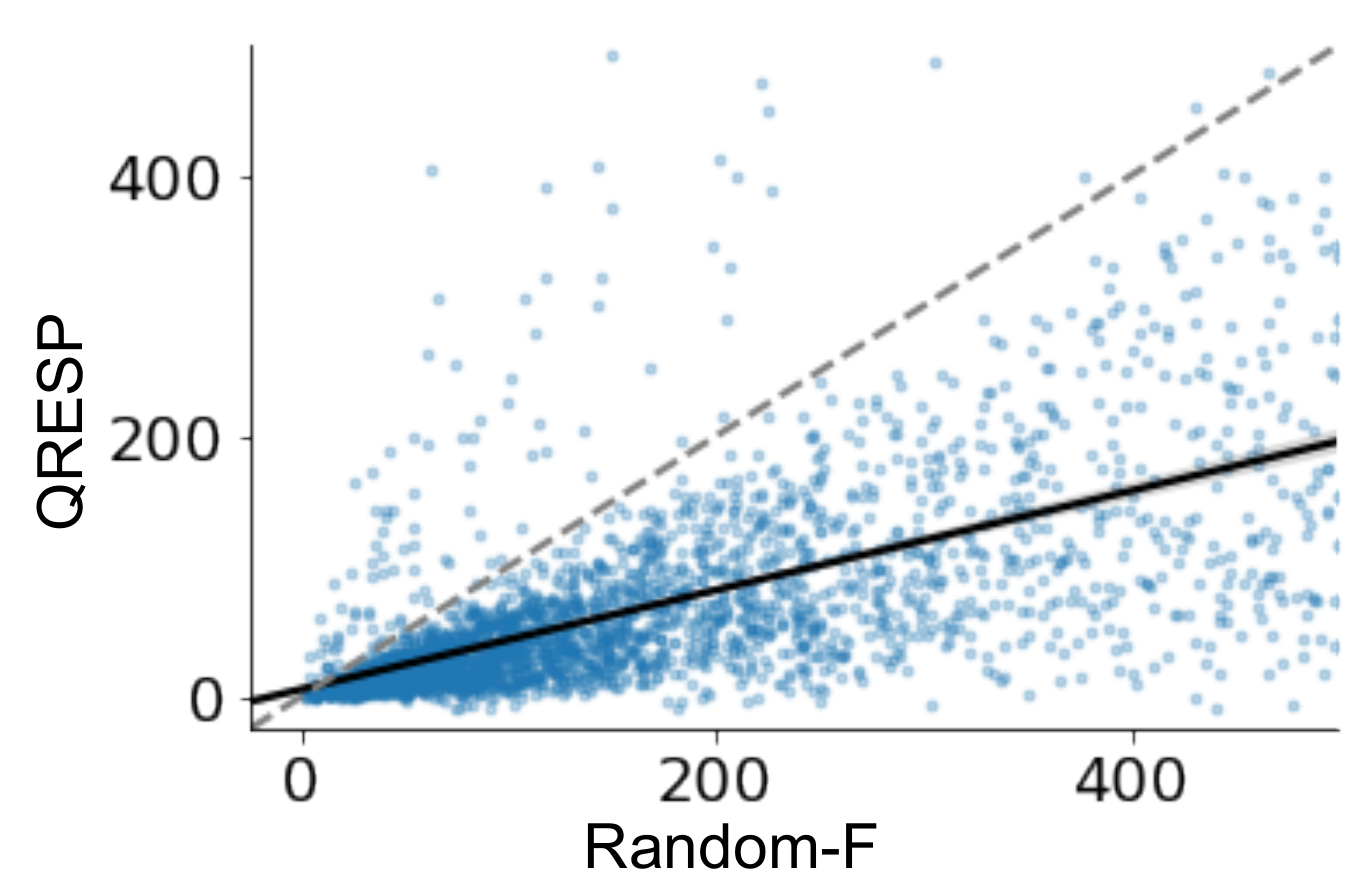}

  \caption{Costs of refinement sets $\Rfilter$ and $\Rsys$. Each point represents a single query $q$. The x-axis indicates the cost $\cS(\Rfilter(q))$ computed by Eq. \ref{eqn:cost}, and the y-axis indicates the cost $\cS(\Rsys(q))$. The solid black line is the least-squares fit. In general, $\Rsys$ refinements achieve much lower cost compared to $\Rfilter$. Points above the diagonal indicate ``search errors'', where the ILP solver reached its time budget before finding a lost-cost solution.}
  \label{fig:gold_scores_scatter_filter}
\end{wrapfigure}

%% file: appx_annotations.tex
\section{Annotation process}  \label{appx:annotation}

Annotations were collected using the Amazon Mechanical Turk platform.

\subsection{Annotators}

Mechanical Turk crowd workers were required to have Masters qualifications, and also required to pass a qualification quiz testing comprehension and good performance on the task. Roughly 25 workers took the qualification quiz; we selected the top 5 to perform annotations.

We maintained contact with annotators via email, fielding questions and discussing challenging cases. Annotators were paid per annotation (or HIT); we chose the HIT rate to target an hourly wage of between \$15 / hour and \$18 / hour.

\subsection{Annotation procedure and annotator agreement}

Each example was annotated by two crowd workers. For the Stage 1 evaluation (see \S \ref{sec:evaluation_criteria}), a refinement is considered fluent if both annotators agree that it is fluent, and similarly for relevance. For Stage 2 evaluation, if one annotator is neutral while the other prefers choice A, we mark choice A as preferred overall. If one annotator prefers A while the other prefers B, we mark the example as neutral.

Table \ref{tbl:kappa} shows measures of inter-annotator agreement as measured by Cohen's $\kappa$, as well as the percentage of refinements that passed Stage 1 quality filters. The $\kappa$ values range between 0.4 and 0.6, often considered to indicate ``moderate'' agreement \cite{Landis1977TheMO}. 

\input{table/kappa_wrapper.tex}

%% file: table/kappa_wrapper.tex

\begin{table}[t]
    \footnotesize
    \centering

    \begin{tabular}{L{3.6em} L{8em} R{5em} R{5em}}
        \toprule
                                          & \textbf{Criterion} & \textbf{Cohen's} $\boldsymbol{\kappa}$ & \textbf{\% passed} \\
        \midrule
        \multirow{2}{*}{\textbf{Stage 1}} & Fluency            & 0.37                                   & 99.1               \\
                                          & Relevance          & 0.61                                   & 90.6               \\
        \midrule
        \multirow{4}{*}{\textbf{Stage 2}}
                                          & Comprehensiveness  & 0.53                                   & -                  \\
                                          & Interestingness    & 0.46                                   & -                  \\
                                          & Non-redundancy     & 0.39                                   & -                  \\
                                          & Overall            & 0.52                                   & -                  \\
        \bottomrule
    \end{tabular}

    \caption{Measures of annotation quality. For Stage 1, ``\% passed'' shows the percentage of refinements passing initial quality filters. The $\kappa$ value for ``Fluency'' is deceptively low, and is a result of the heavily imbalanced label distribution: $> 99\%$ of refinements passed the fluency screen.}
    \label{tbl:kappa}
\end{table}

%% file: appx_stats.tex
\section{Statistical testing}  \label{appx:stats}

For the human evaluations reported in \S \ref{sec:human_eval_gold} and \S \ref{sec:eval_human}, we perform statistical tests to determine whether system A (\sysname) is preferred by annotators over system B at a rate statistically different from chance.

For Stage 1, we test the null hypothesis that refinements from systems A and B pass the Stage 1 screen at the same rate. We construct a contingency table where the first column contains counts for the number of refinements from system A that pass and fail the screen, respectively. The second column contains similar counts for system B. We then perform Fisher's exact test\footnote{\url{docs.scipy.org/doc/scipy/reference/generated/scipy.stats.fisher_exact.html}} on the resulting contingency table.

For Stage 2, we perform a separate test for each of the four attributes. In particular, we perform a binomial test\footnote{\url{docs.scipy.org/doc/scipy/reference/generated/scipy.stats.binomtest.html}} to test the null hypothesis that a rater is equally likely to prefer A as to prefer B, against the alternative that a rater is more likely to prefer A. For simplicity, we excluded ``neutral'' cases where the rater could not decide between A and B.

%% file: appx_generation.tex
\section{Refinement generation}  \label{appx:generation}

\subsection{Selection of queries for \nqtrec} \label{appx:nqtrec_selection}

In the interest of evaluating our models on real-world list-intent queries, we selected a handful of evaluation queries from Natural Questions and from the TREC 2009 Million Query track. For Natural Questions, we filtered down to questions with answers found in a table or list, and then manually selected a collection of queries from this list. We aimed to select list-intent queries with a large number of potential answer entities, across a variety of domains. For TREC, we similarly filtered down to queries from the \texttt{List} query class, and then manually selected queries for diversity and interest. The TREC data can be downloaded from \url{https://trec.nist.gov/data/million.query09.html}. 

The full list of TREC and NQ evaluation queries is included in Appendix \ref{appx:eval_queries}.

\subsection{Additional human evaluations} \label{appx:human_eval}

In \S \ref{sec:eval_human}, we compare the results of $\Msys$ against $\Mrandom$ and $\Mseparate$ on queries from YAGO. We also performed A / B tests for $\Mfilter$; the results are generally similar to what we found comparing against $\Mrandom$ and are shown in Table \ref{tbl:ab_yago_pred}. $\Msys$ shows a smaller advantage in interestingness over $\Mfilter$ compared to $\Mrandom$; this likely occurs since $\Mfilter$ is not trained on generic refinements which simply add locations or dates.

\input{table/ab_yago_pred_appx_wrapper.tex}

%% file: table/ab_yago_pred_appx_wrapper.tex

\begin{table}[h!]
    \footnotesize
    \centering

    \input{table/human_evals/ab-yago-filter_joint_best_gold_false-filter_joint_random_gold_false.tex}

    \caption{Results of A / B tests on the YAGO human evaluation set, comparing $\Msys$ against $\Mfilter$. $\Msys$ is more comprehensive. $^{**}$ indicates $p < 0.001$.}
    \label{tbl:ab_yago_pred_appx}
\end{table}

%% file: table/human_evals/ab-yago-filter_joint_best_gold_false-filter_joint_random_gold_false.tex
\begin{tabular}{
    L{6.7em} *{3}{L{3.4em}}
  }
  \toprule
                                 & \multicolumn{3}{c}{\textbf{A} = \sysname vs. \textbf{B} = \filtername} \\
  \cmidrule(lr){2-4}
  $N = 107$                      & \textbf{A}              & Neutral     & \textbf{B}                     \\
  \midrule
  \tworows{Fluent + \\ Relevant} & \tworows{\textbf{94\%}} & \tworows{-} & \tworows{92\%}                 \\
  \\
  \cmidrule(lr){1-4}
  Comprehensive                  & \textbf{59\%}$^{**}$    & 21\%        & 20\%                           \\
  Interesting                    & \textbf{39\%}           & 35\%        & 26\%                           \\
  Non-redundant                  & \textbf{23\%}           & 62\%        & 15\%                           \\
  Overall                        & \textbf{59\%}$^{**}$    & 23\%        & 18\%                           \\
  \bottomrule
\end{tabular}

%% file: appx_eval_queries.tex
\section{Evaluation queries} \label{appx:eval_queries}

The full list of human evaluation queries for YAGO and \nqtrec is shown in Table \ref{tbl:eval_queries}.

\input{table/eval_queries.tex}

%% file: table/eval_queries.tex
\begin{table*}[t]
    \footnotesize
    \centering
    \begin{subtable}[h]{\linewidth}
        \begin{tabular}{
                L{46em}
            }
            \toprule
            Academic journals, Action films, Activists, Albums, American activists, American artists, American military officers, American songs, Animals, Aviators, Baseball players, Battles, Battles of the Middle Ages, Birds, Boxers, Brazilian people, Bridges, British musicians, Buildings and structures in England, Canadian people, Charitable organizations, Chinese people, Christian saints, Classical musicians, Comics characters, Companies based in Tokyo, Companies of the United States, Concertos, Dance music albums, Dictionaries, Electronic albums, Energy companies, Engineers, Engines, European films, Films, Finnish writers, Firearms, Foods, French novels, German films, History books, History museums, Indian films, Indian writers, Insect genera, Islands, Italian painters, Japanese songs, Lakes, Languages, Locomotives, Magazines, Manufacturing companies, Martial arts people, Media executives, Medical journals, Minerals, Mobile phones, Music award winners, Musicians, Newspapers, Non-fiction books, Novels, Operas, Organisations based in India, Organisations based in Singapore, Organizations based in the United States, Painters, Pakistani films, People associated with crime, People associated with religion, People from New York City, People in finance, Philosophers, Philosophical works, Physicians, Plays, Poets, Political organizations, Political parties, Proteins, Racing drivers, Radio stations, Researchers, Rock songs, Schools in London, Scientists, Ships, Singers, Social scientists, Songs, Sports competitions, Sports events, Sports leagues, Sports venues, Swimmers, Tools, Typefaces, United States federal judges, Vehicles, Video games, Weapons, Websites, Writers \\
            \bottomrule
            \caption{YAGO queries}
        \end{tabular}
    \end{subtable}
    \begin{subtable}[h]{\linewidth}
        \begin{tabular}{
                L{46em}
            }
            \toprule
            19th-century artists, Active volcanoes in the Philippines, African wool producers, Apple products, Architects in New Jersey, Arguments for the existence of God, Astronauts who stepped on the moon, BBC science news, BMW car models, Backward compatible games for XBox One, Baseball players featured on postage stamps, Battles of the revolutionary war, Best selling artists of all time, Billboard hot 100 number-one singles, Books in the New Testament, Branches of medicine, Bright stars in the sky, Cast of the movie ``Now you see me'', Causes of the French Revolution, Census regions in the United States, Chief ministers of Indian states, Cities and towns in Northern California, Cities that have held the Olympic Games, Cities with high murder rates, Communist countries during the Cold War, Countries where US citizens can travel without a visa, Countries with French-speaking people, Democratic countries, Disney Pixar movies, Disney princesses, Division 2 colleges in the midwest, Earthquakes, Foods brought to the New World from Europe, Forbes list of largest companies in the world, Functions of all the body systems, Functions of the government, Games for super nintendo classic, Gods and goddesses of the world, Government monopolies in the United States, Greatest NBA players of all time, Hall of fame football players, Hotels near downtown Houston, Independent power producers in South Africa, Indian spices, James Bond movies, Jesuit universities in the United States, John D Rockefeller's philanthropic projects, Languages spoken in India, Large charities, Largest cities in the world, Major exports of the United States, Malayalam movies, Most densely populated areas in the world, Most frequently used words in English, Most popular video games, Most spoken languages in the world, Movies Robert de Niro played in, National monuments in the United States, Natural air pollutants, Natural resources, Nobel Prize winners, Oil and gas companies in Kuwait, Oscar winners, PS4 games, Participants at the Battle of Wounded Knee, Places where carbon is stored on earth, Players who have a receiving touchdown in a superbowl, Political parties in India, Popular YouTube channels, Private medical colleges in Sindh, Public high schools in Brooklyn New York, Public sector mutual funds in India, Renewable energy companies, Rock and Roll Hall of Fame artists, Roles of local government in the Philippines, Romantic anime shows in English dub, Rulers of England, Satellites launched by India, Schools that offer architecture in Nigeria, Songs in West Side Story, Songs with California in the title, Sources of US oil, States in Nigeria, Stocks in the Dow Jones industrial average, Supreme court justices, Time Magazine person of the year winners, Trees with heart-shaped leaves, US presidents, Universities and colleges in Australia, Walt Disney films, World stock exchanges, Wrestling promotions in the United States, XBox 360 games \\
            \bottomrule
            \caption{\nqtrec queries}
        \end{tabular}
    \end{subtable}

    \caption{Full list of queries used for human evaluation.}
    \label{tbl:eval_queries}
\end{table*}

%% file: appx_annotation_guide.tex
\section{Annotation guide} \label{appx:annotation_guide}

The annotation UI is shown in Fig. \ref{fig:annotation_ui}. The instructions summary is show in Fig. \ref{fig:annotation_summary}. Detailed instructions are shown in Fig. \ref{fig:annotation_guide_details}.

\clearpage

\input{fig/annotation_guide_wrapper.tex}

%% file: fig/annotation_guide_wrapper.tex
\begin{figure*}[t]
  \centering
  \includegraphics[width=\textwidth]{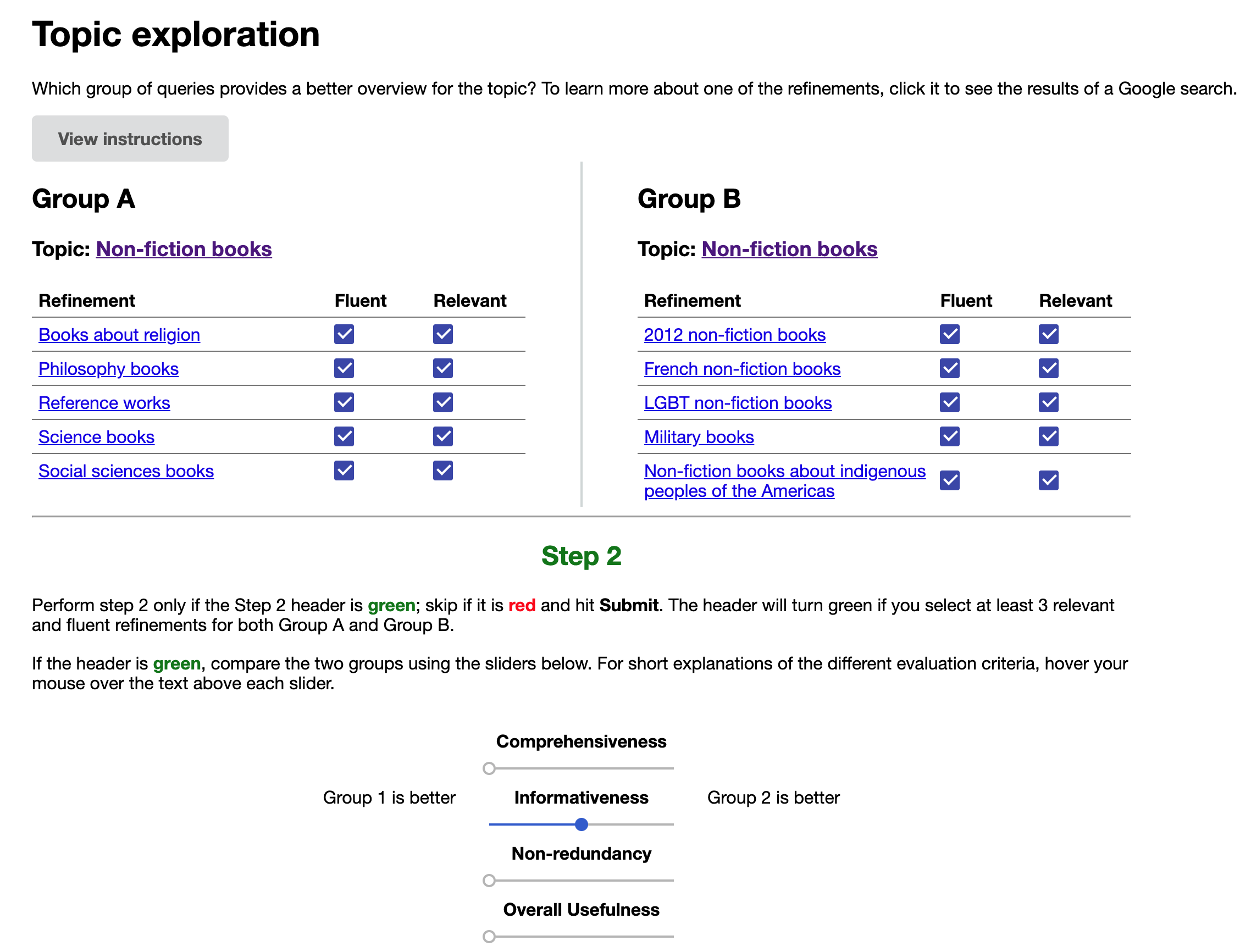}

  \caption{Annotation UI and summary of annotation instructions.}
  \label{fig:annotation_ui}
\end{figure*}

\begin{figure*}[t]
  \centering
  \includegraphics[width=\textwidth]{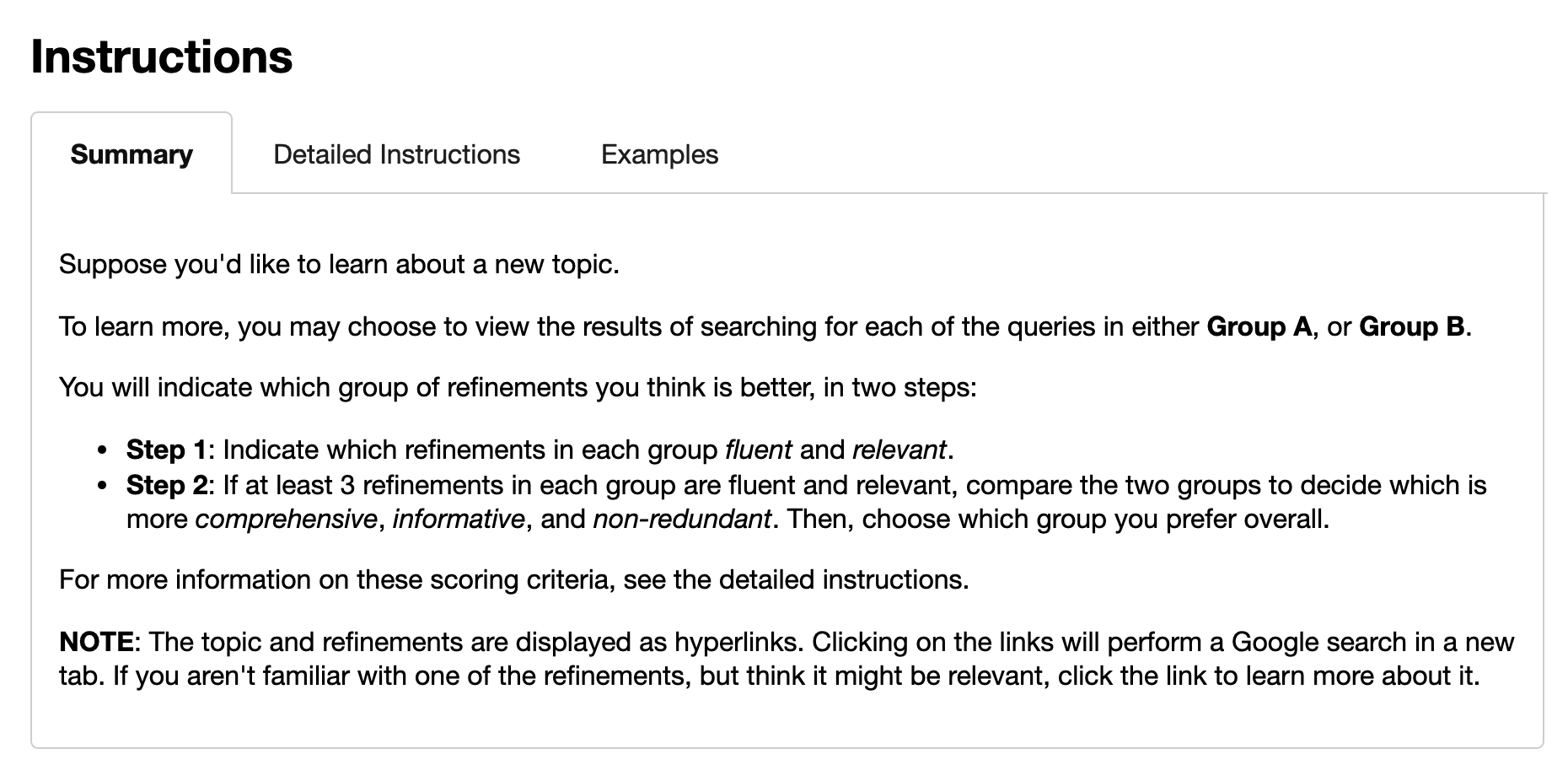}

  \caption{Annotation UI and summary of annotation instructions.}
  \label{fig:annotation_summary}
\end{figure*}

\begin{figure*}[t]
  \centering
  \includegraphics[width=0.9\textwidth]{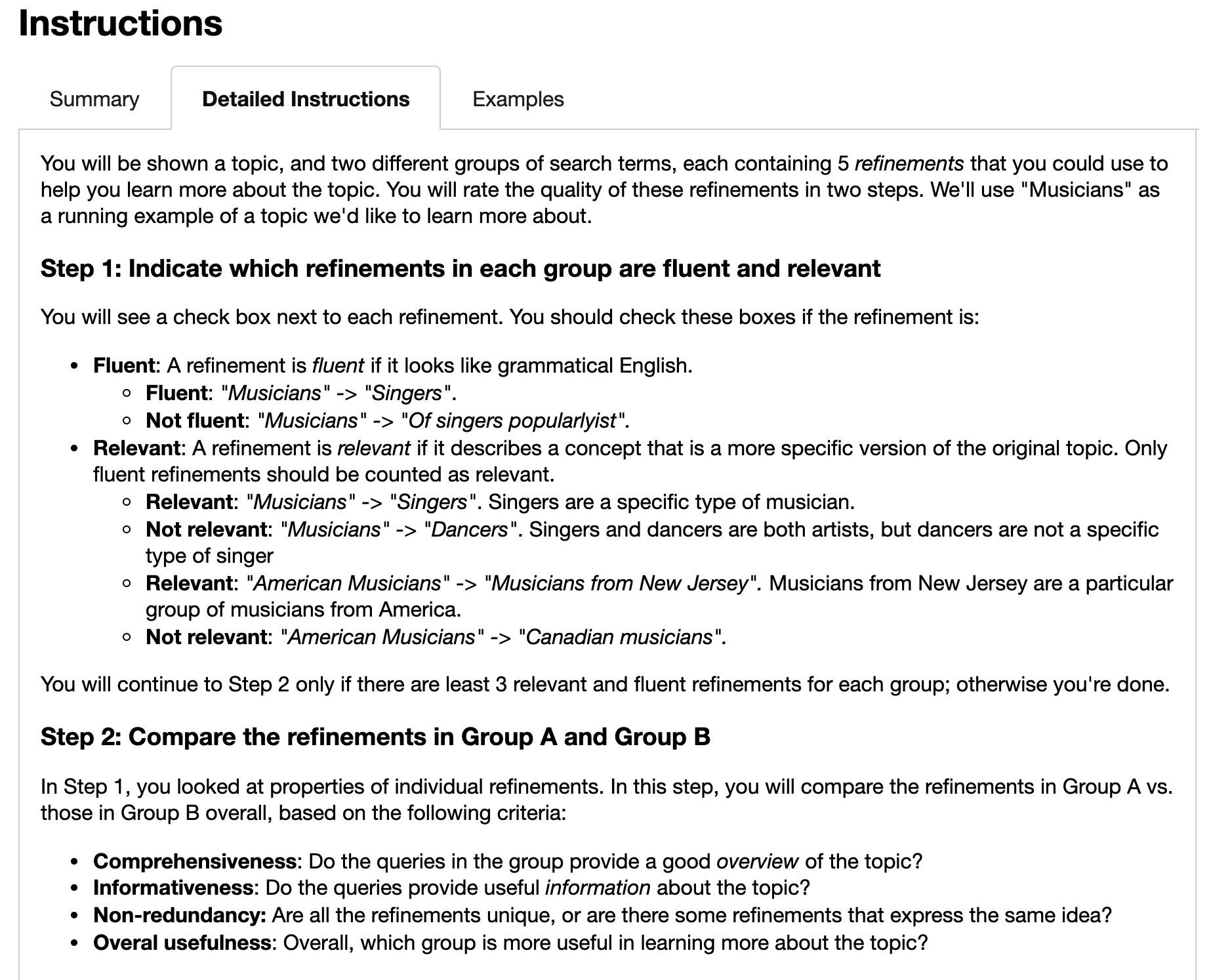}
  \includegraphics[width=0.9\textwidth]{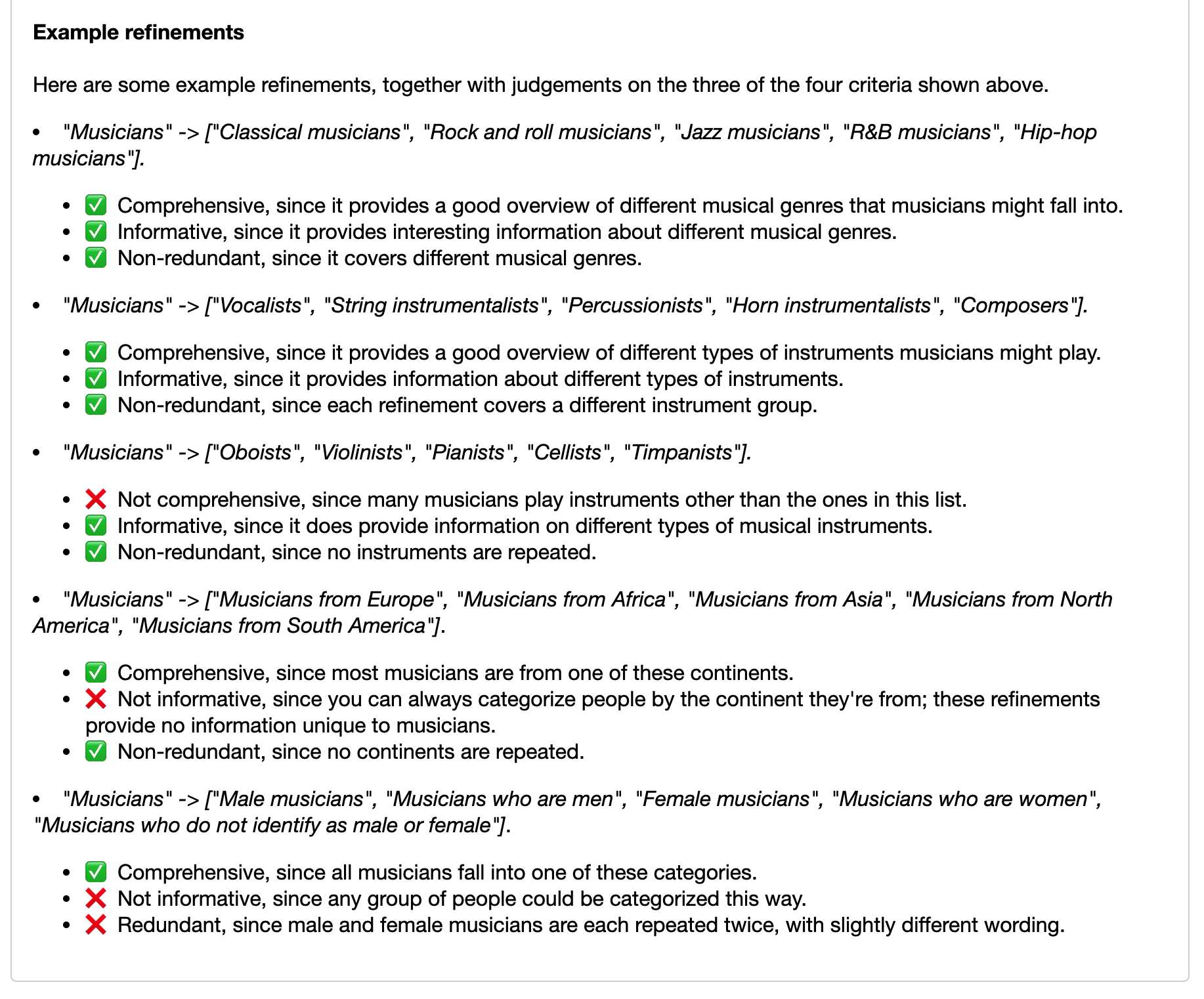}

  \caption{Annotation UI and summary of annotation instructions.}
  \label{fig:annotation_guide_details}
\end{figure*}